\documentclass[sigconf]{acmart}

\usepackage{multirow}
\usepackage{makecell}
\usepackage{enumitem}
\usepackage{longtable}

\AtBeginDocument{%
  }

\copyrightyear{2024}
\acmYear{2024}
\setcopyright{acmlicensed}
\acmConference[KDD '24] {Proceedings of the 30th ACM SIGKDD Conference on Knowledge Discovery and Data Mining }{August 25--29, 2024}{Barcelona, Spain.}
\acmBooktitle{Proceedings of the 30th ACM SIGKDD Conference on Knowledge Discovery and Data Mining (KDD '24), August 25--29, 2024, Barcelona, Spain}
\acmISBN{979-8-4007-0490-1/24/08}
\acmDOI{10.1145/XXXXXX.XXXXXX}
\begin{document}

\title{Leveraging Pedagogical Theories to Understand Student Learning Process with Graph-based Reasonable Knowledge Tracing}

\author{Jiajun Cui}
\orcid{0000-0001-5900-7643}
\email{cuijj96@gmail.com}
\affiliation{%
  \institution{East China Normal University}
  \city{Shanghai}
  \country{China}
}

\author{Hong Qian}
\orcid{0000-0003-2170-5264}
\email{hqian@cs.ecnu.edu.cn}
\affiliation{%
  \institution{East China Normal University}
  \city{Shanghai}
  \country{China}
}

\author{Bo Jiang}
\orcid{0000-0002-7914-1978}
\email{bjiang@deit.ecnu.edu.cn}
\affiliation{%
  \institution{East China Normal University}
  \city{Shanghai}
  \country{China}
}

\author{Wei Zhang}
\orcid{0000-0001-6763-8146}
\authornote{Corresponding author. This work was supported in part by National Key R\&D Program of China (No. 2023YFC3341200), National Natural Science Foundation of China (No. 92270119 and No. 62072182), and Key Laboratory of Advanced Theory and Application in Statistics and Data Science, Ministry of Education.}
\email{zhangwei.thu2011@gmail.com}
\affiliation{%
  \institution{East China Normal University}
  \city{Shanghai}
  \country{China}
}

\renewcommand{\shortauthors}{Cui et al.}

\begin{abstract}
Knowledge tracing (KT) is a crucial task in intelligent education, focusing on predicting students' performance on given questions to trace their evolving knowledge.
The advancement of deep learning in this field has led to deep-learning knowledge tracing (DLKT) models that prioritize high predictive accuracy.
However, many existing DLKT methods overlook the fundamental goal of tracking students' dynamical knowledge mastery.
These models do not explicitly model knowledge mastery tracing processes or yield unreasonable results that educators find difficulty to comprehend and apply in real teaching scenarios.
In response, our research conducts a preliminary analysis of mainstream KT approaches to highlight and explain such unreasonableness.
We introduce GRKT, a graph-based reasonable knowledge tracing method to address these issues.
By leveraging graph neural networks, our approach delves into the mutual influences of knowledge concepts, offering a more accurate representation of how the knowledge mastery evolves throughout the learning process. Additionally, we propose a fine-grained and psychological three-stage modeling process as knowledge retrieval, memory strengthening, and knowledge learning/forgetting, to conduct a more reasonable knowledge tracing process.
Comprehensive experiments demonstrate that GRKT outperforms eleven baselines across three datasets, not only enhancing predictive accuracy but also generating more reasonable knowledge tracing results. This makes our model a promising advancement for practical implementation in educational settings. The source code is available at \url{https://github.com/JJCui96/GRKT}.
\end{abstract}

\renewcommand{\shorttitle}{Graph-based Reasonable Knowledge Tracing}



\maketitle

\section{Introduction}\label{sec:intro}

In personalized learning, Knowledge Tracing (KT) is crucial for tracking students' evolving knowledge mastery based on their historical question responses~\cite{wu2024comprehensive,corbett1994knowledge}.
Early researchers addressed this challenge by leveraging the monotonicity assumption~\cite{embretson2013item}, linking better mastery of one knowledge concept (KC) to a higher probability of correctly answering related questions.
They trained models to predict student responses on given questions, proposing typical machine learning-based KT methods~\cite{corbett1994knowledge,pardos2011kt}. Consequently, predicting student performance became the primary task, with prediction accuracy as the mainstream metric for evaluating KT models, promoting the emergence of deep learning knowledge tracing  (DLKT) methods. However, many DLKT approaches prioritize prediction ability over the fundamental objective of knowledge tracing, sometimes forgoing tracing altogether~\cite{choi2020towards, ghosh2020context}. Others use internal network weights to represent knowledge mastery~\cite{yin2023tracing, shen2021learning}, facing challenges in constructing meaningful tracing results due to the low interpretability and reasonability of deep neural network structures.
Hidden neurons in these networks adaptively learn from data without explicit meaning~\cite{guidotti2018survey}.
It is worth noting that the cognitive diagnosis task also assesses knowledge mastery but usually focuses on static testing instead of dynamic learning process~\cite{liu2021survey,leighton2007cognitive}.
Therefore, we do not delve into it within this paper.

\begin{figure*}[!t]
  \centering
  \includegraphics[width=1.0\linewidth]{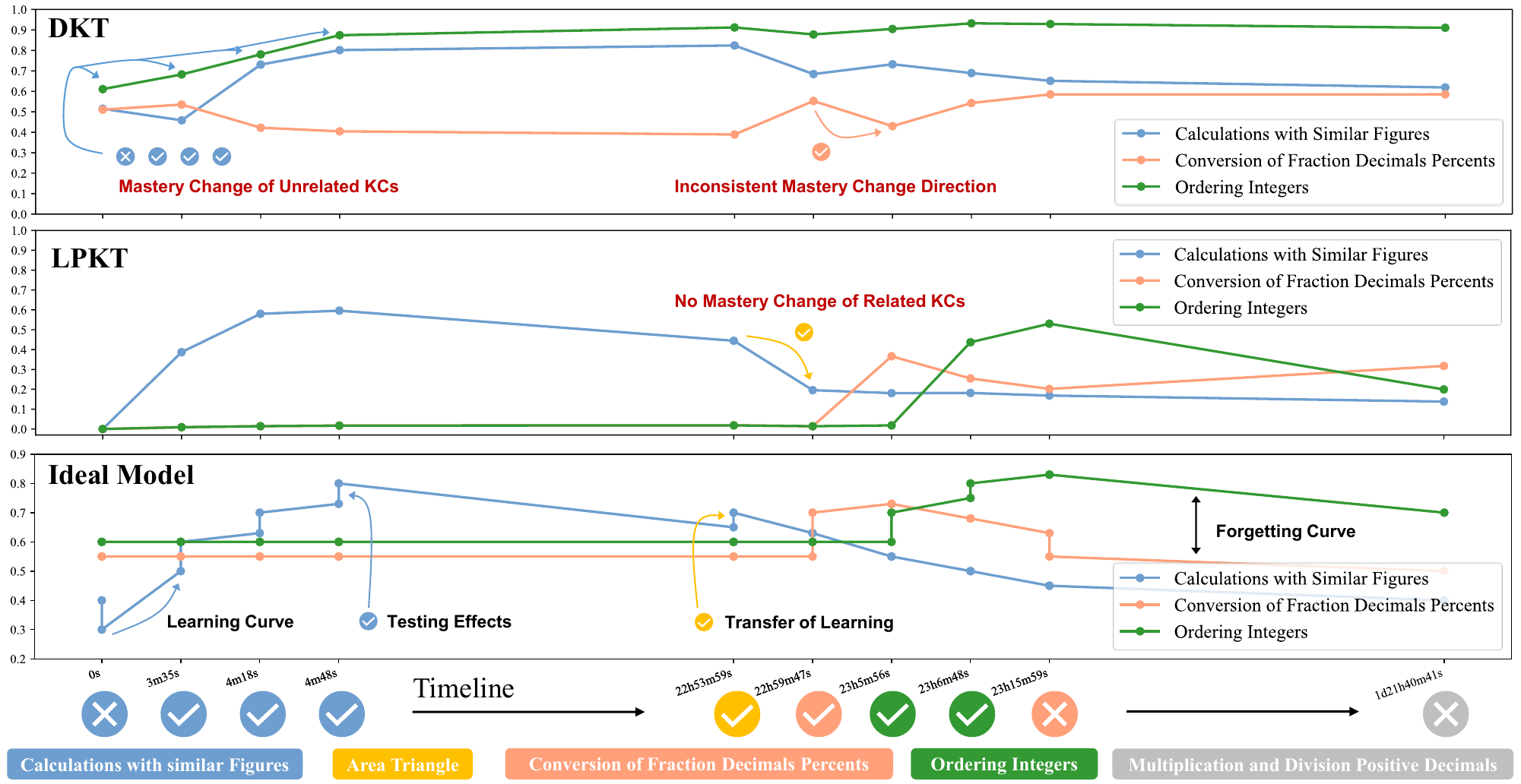}
  \caption{Illustration of a student's evolving knowledge mastery while answering ten questions, traced by two DLKT models, along with an assumed ideal tracing result. The student is sampled from the ASSIST12 dataset, introduced in Section~\ref{sec:dataset}.}
 \label{fig:intro}
 \vspace{-0.5em}
\end{figure*}

Figure~\ref{fig:intro} illustrates the traced dynamic knowledge mastery of an example student by two DLKT models: DKT~\cite{piech2015deep} and LPKT~\cite{shen2021learning}.
DKT is a pioneering approach that directly applies recurrent neural networks (RNNs) to the KT task.
In this case, when the student responds to the initial four questions related to the blue KC \textit{Calculations with Similar Figures}, their knowledge mastery of the unrelated green KC \textit{Ordering Integers} increases, presenting an unreasonable outcome.
Furthermore, a correct response to the sixth question results in a contrary decrease in its corresponding KC's mastery, demonstrating an inconsistent change in direction.
LPKT, as a time-aware method, models learning and forgetting processes for more reasonable knowledge tracing.
However, it struggles to capture the relation between the yellow KC \textit{Area Triangle} and the blue KC \textit{Calculations with Similar Figures}, as evidenced by the decreasing mastery of the blue curve following a correct response to an question of yellow.
Both of these two KCs examine students' calculations about the base and height of triangles, which suggests their underlying relation.
Beneath the figure is a tracing result from an assumed ideal model, which we design based on comprehensive pedagogical effects.
As shown, the student mastery will increase and drop according to their right/wrong responses based on the testing effect~\cite{roediger2006test}.
The mastery of the yellow KC would relatedly increase due to the correct response to the sixth orange KC, according to the transfer of learning~\cite{perkins1992transfer}.
Besides, the mastery between responses should also vary due to students' learning and forgetting behaviors modeled by the learning and forgetting curves~\cite{yelle1979learning, ebbinghaus1885gedachtnis}.

From this example, we summarize three deficiencies of current DLKT methods in dynamic knowledge tracing reasonability: (i) \textbf{Mastery change of unrelated KCs} - learning one KC affects unrelated KC mastery; (ii) \textbf{No mastery change of related KCs} - learning one KC does not impact related KCs; (iii) \textbf{Inconsistent mastery change direction} - correct answers may decrease KC mastery, and vice versa.
These stem from opaque deep neural networks, whose parameters serve the overarching objective of performance prediction.
Moreover, many researches use RNNs to model knowledge application and update by the recurrent units' output and state transition~\cite{shen2021learning,shen2022assessing,liu2019ekt,piech2015deep}.
This mixes the effects of students answering questions and their spontaneous behaviors, leading to confusing tracing results.
For example, incorrect responses may strengthen wrong knowledge retrieval and get a mastery drop of the related KC.
But when they get feedback and learn from their errors, they can make a final progress.
This fine-grained knowledge mastery changing is not captured.
To address these above issues, we introduce GRKT, a \textbf{G}raph-based \textbf{R}easonable \textbf{K}nowledge \textbf{T}racing to enhance knowledge tracing reasonability while retaining neural networks' representational power.

To be specific, we integrate pedagogical theories~\cite{perkins1992transfer,yelle1979learning,ebbinghaus1885gedachtnis,roediger2006test} into the KT modeling, dividing the learning process into three distinct stages.
(i) The \textbf{knowledge retrieval} stage analyzes how students respond to questions.
This stage draws from cognitive psychology~\cite{melton1963implications}, viewing learning as encoding, storing, and retrieving memories.
When students answer questions, retrieval from memory becomes crucial.
We start this stage by retrieving the encoded memory related to the question's KC and project it into a mastery value.
We then compare this value with the question's difficulty score to predict if the student could correctly answer the question.
(ii) The \textbf{memory strengthening} stage focuses on how answering questions impacts students' knowledge mastery.
Here, students strengthen their memory retrieval routes, aligning with the \textit{Testing Effect} theory~\cite{roediger2006test, kornell2009unsuccessful}.
Correct retrievals enhance learning, while incorrect ones reinforce errors.
We encode this positive/negative memory strengthening in the knowledge memory of the relevant KC based on whether the question is correctly solved.
(iii) The \textbf{knowledge learning/forgetting} stage explores what students do after question answering.
This stage aims to model the active learning and natural forgetting behaviors based on the \textit{Learning curve}~\cite{yelle1979learning} and the \textit{Forgetting curve}~\cite{ebbinghaus1885gedachtnis}.
Both curves suggest a decreasing rate of learning and forgetting over time.
Concretely, we first introduce a learning decider to determine whether students will continue learning the KCs just practiced or the KCs for future study.
Then, we employ KC-specific time-aware kernels to model the learning/forgetting curves of all involved KCs based on these decisions.
By applying this three-stage modeling process iteratively across students' response sequences, we establish a coherent and reasonable knowledge tracing framework.
This approach effectively captures mastery changes resulting from question answering and subsequent behaviors, addressing the issue of \textbf{inconsistent mastery change direction}.

To handle the two other issues of \textbf{mastery change of unrelated KCs} and \textbf{no mastery change of related KCs}, we utilize the message passing mechanism of graph neural networks (GNNs) applied to KC relation graphs. 
This mechanism establishes clear boundaries between related and unrelated KCs. 
Specifically, changes in knowledge mastery of one KC are propagated through the graph edges to its related KCs within a specific number of hops. 
From the pedagogical perspective, this message passing aligns with the \textit{Transfer of Learning theory}~\cite{perkins1992transfer}, which explains humans' ability to transfer knowledge between similar fields to solve problems and acquire skills. 
We integrate this understanding into our three-stage learning process modeling using KC relation-based GNNs. 
For instance, in the first stage of GRKT, instead of solely retrieving knowledge from the target question's KC, we utilize graph aggregation to synthesize the memory of the KC's neighbors for solving the question. 
Similarly, during the second stage, the memory strengthening process involves propagating the gain and loss of knowledge mastery to the KC's neighbors, and this process is also applied in the third stage's knowledge learning.
Additionally, we exploit the homophily of GNNs to generate similar time-aware kernels for related KCs, effectively modeling their similar learning/forgetting processes. 
This defines the boundaries between related and unrelated KCs based on the number of hops in GNN operations, effectively addressing challenges associated with mastery changes between different KCs. 
It's worth noting that KCs have various types of relations, including prerequisite, similarity, collaboration, remedial, and hierarchy~\cite{gao2023leveraging}. 
In GRKT, we primarily focus on leveraging the two most commonly used relations: prerequisite and similarity.

To the best of our knowledge, this work represents the first comprehensive analysis of the reasonability issues in current DLKT methods, and integrates multiple pedagogical theories to address these concerns.
The main contributions of this paper are as follows:

\begin{itemize}[leftmargin=*]
\item \textbf{Motivation.}
We identify the reasonability issues arising from the widespread adoption of deep learning techniques in the KT task.
Many DLKT methods tend to excessively prioritize student performance prediction, often overlooking unreasonable knowledge tracing results due to the inherent interpretability challenges posed by neural networks.
\item \textbf{Methods.}
We outline three primary reasonability issues prevalent in current DLKT methods. To address these issues, we introduce GRKT, a graph-based reasonable knowledge tracing, which establishes a three-stage learning process modeling.
Additionally, we utilize the KC relation graph to mitigate mutual effects among KCs.
The incorporation of multiple pedagogical theories provide sufficient support for our proposed method.

\item\textbf{Experiments.}
Comprehensive experimental results showcase that our GRKT exhibits superior prediction performance and yields reasonable knowledge tracing results when compared to eleven baselines across three widely-used datasets.
\end{itemize}

\section{related work}

\subsection{Reasonable Knowledge Tracing}
Early KT methods in machine learning, such as Bayesian Knowledge Tracing (BKT)~\cite{corbett1994knowledge}, initially showcased reasonable results due to their transparent and interpretable internal structure.
BKT utilizes Hidden Markov Models (HMMs) to probabilistically represent the student learning process.
It transitions knowledge mastery and emits probabilities of correct responses, while also considering guessing and slipping behaviors.
Subsequent KT methods expanded upon BKT by incorporating additional pedagogical factors such as question difficulty~\cite{pardos2011kt} or prior student information~\cite{yudelson2013individualized}.

However, traditional methods show inferior prediction performance when compared to subsequent emerging DLKT methods~\cite{piech2015deep, liu2019ekt, shen2022assessing,pandey2019self, ghosh2020context, choi2020towards, cui2024interpretable},
which reach high prediction performance due to the power of neural networks.
Even so, these DLKT methods fail to produce reasonable knowledge tracing results due to their inherently opaque structures. Efforts have been made to tackle this challenge.
Shen et al.~\cite{shen2021learning} proposed Learning Process-consistent Knowledge Tracing (LPKT), which utilizes student response duration and interval time to capture learning and forgetting behaviors. However, it only focuses on knowledge learning and forgetting and does not model the interplay of knowledge mastery changes between KCs, limiting its reasonability.
Similarly, Yin et al.~\cite{yin2023tracing} introduced the Diagnostic Transformer (DTransformer), which diagnoses student knowledge mastery from each tackled question and employs a contrastive learning framework to produce more stable knowledge tracing. While this stability enhances reasonability to some extent, its transformer-based structures do not adequately reflect the transition of knowledge mastery between continuous student responses.
Therefore, while these approaches improve model reasonability from specific angles, they do not offer a comprehensive method to generate reasonable knowledge tracing results covering both KC relations and continuous learning processes.

We address this gap with our proposed GRKT, which utilizes GNNs to model KC relations and introduces a three-stage learning process to capture evolving knowledge mastery.
By integrating these techniques, GRKT achieves high prediction performance while also generating more reasonable knowledge tracing results.

\subsection{Graph-based Knowledge Tracing}
Graph Neural Networks (GNNs)~\cite{scarselli2008graph} serve as an efficient tool to capture intricate relations between instances in real-world scenarios.
Their message aggregation and propagation operations on graphs yield deep representations for node features, enhancing performance in various downstream tasks across different domains.
In the context of KT, researchers explore various structures to harness the power of GNNs.
Nakagawa et al.~\cite{nakagawa2019graph} pioneered the incorporation of GNNs into KT by reformulating it as a time-series node-level classification problem based on KC relation graphs.
Gan et al.~\cite{gan2022knowledge} leveraged this structure to enhance graph representation learning, generating more informative question and concept embeddings.
Except for KC relations, question-question and question-KC relations are also widely considered.
For instance, Bi-CLKT~\cite{song2022bi} applied contrastive learning to question-KC and KC-KC graphs to generate question embeddings enriched with question and KC structural information.
Another work~\cite{yang2021gikt} leveraged question-KC relations to address question sparsity and multi-skill problems.
In our paper, we specifically focus on utilizing GNNs to model the mutual effects of KCs during students' knowledge leveraging and changing, constructing a more reasonable approach to knowledge tracing.

It is worth noting that some other GNN-based or memory-based methods (e.g., GKT~\cite{nakagawa2019graph} and DKVMN~\cite{zhang2017dynamic}) also update mastery between KCs.
However, their knowledge state updating is still potentially performed by the erase-followed-by-add mechanism, which uses GRU/LSTM cells unable to solve the reasonability issues such as not guaranteeing the direction of consistency change between KCs.

\newcommand{\Hset}{\mathcal{H}}
\newcommand{\Uset}{\mathcal{U}}
\newcommand{\Qset}{\mathcal{Q}}
\newcommand{\Cset}{\mathcal{C}}
\newcommand{\Mset}{\mathcal{M}}
\newcommand{\Nset}{\mathcal{N}}

\section{preliminary}
\subsection{Task Formulation}

Knowledge tracing aims to trace the dynamic evolution of students' knowledge mastery throughout their learning processes characterized by their responses to questions. 
Suppose there are a student set $\Uset$, a question set $\Qset$, and a KC set $\Cset$.
Each student $u\in\Uset$ has a historical response sequence $\Hset^u=\{r^u_1,r^u_2,\cdots,r^u_{|\Hset^u|}\}$, where each response $r^u_t=\left(q^u_t,a^u_t,c^u_t,T^u_t\right)$ comprises the involved question $q^u_t\in\Qset$, the correctness $a^u_t\in\{0, 1\}$ (where $a^u_t=1$ means a correct response), the KC $c_t^u\in\Cset$ examined by the question, and the timestamp $T_t^u$ of the response.
It is worth noting that there could be multiple KCs associated with one question.
To be concise, we use the notations with just one KC to describe the task setting and the proposed method, but our method is easily extended to the setting of multiple KCs (e.g., averaging the KC representations as mentioned in Section~\ref{subsec:kr}).
The objective is to track and monitor the evolving knowledge mastery of $u$ after each response, $\Mset^u=\{\textbf{m}^u_1,\textbf{m}^u_2,\cdots,\textbf{m}^u_{|\Hset^u|}\}$ where $\textbf{m}^u_t$ is stacked with $\{m^u_{c_i,t}|c_i\in\Cset\}$ and $m^u_{c_i,t}$ signifies the student's knowledge mastery of the KC $c_i$ at time step $t$.
A higher value denotes a superior level of mastery.
However, the absence of annotated mastery levels necessitates researchers to resort to the student performance prediction task as a surrogate measure~\cite{liu2021survey}.
In this paradigm, given $\Hset^u$, the objective is to predict whether student $u$ can correctly answer a new question $q^u_{|\Hset^u|+1}$, with its associated KC $c^u_{|\Hset^u|+1}$ at timestamp $T^u_{|\Hset^u|+1}$.
This hinges on the monotonicity assumption~\cite{embretson2013item}, which posits that higher knowledge mastery leads to a higher probability of answering questions correctly.
For brevity, we omit the superscript $u$ in the later method description.

\newcommand{\Gg}{\mathcal{G}}
\newcommand{\Pg}{\mathcal{P}}
\newcommand{\Sg}{\mathcal{S}}
\newcommand{\Rg}{\mathcal{R}}
\newcommand{\Gset}{\Gg\in\{\Pg,\Sg,\Rg\}}
\newcommand{\real}[2]{\mathbb{R}^{#1\times #2}}
\newcommand{\trans}[1]{{#1}^{\text{T}}}

\begin{figure*}[!t]
  \centering
  \includegraphics[width=\linewidth]{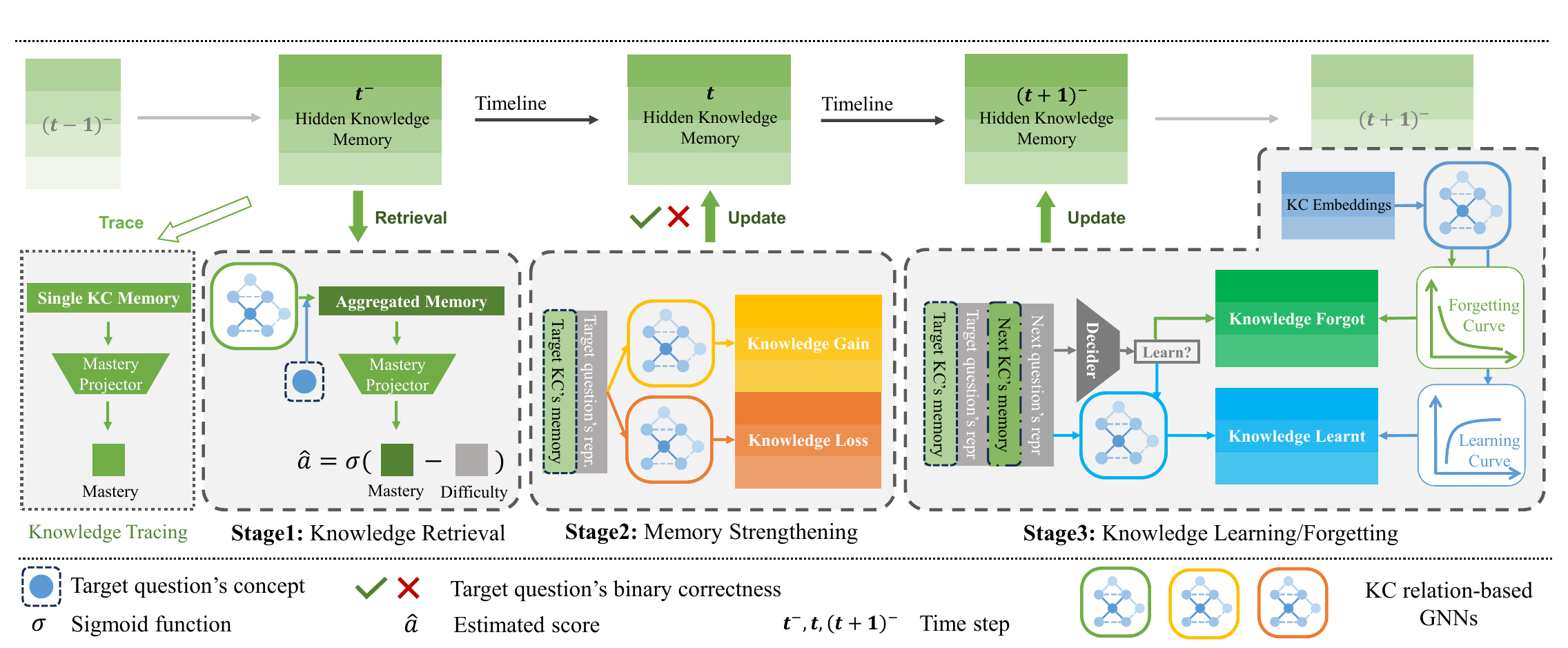}
  \caption{The entire framework of GRKT encompasses three recurrent stages: knowledge retrieval, memory strengthening, and knowledge learning/forgetting.}
 \label{fig:overview}
\end{figure*}

\section{Methodology}
As shown in Figure~\ref{fig:overview}, GRKT conducts a recurrent modeling within a three-stage learning process: knowledge retrieval, memory strengthening, and knowledge learning/forgetting.
The proposed KC relation-based graph neural networks capture knowledge mastery variation between KCs throughout these stages.
This section introduces the KC relation-based GNNs first, then explains the three-stage learning process modeling with these GNNs.
For ease of understanding GRKT, we list and explain all relevant notations in Appendix~\ref{ap:notation}.

\subsection{KC Relation-based Graph Neural Networks}
Based on the transfer of learning theory~\cite{perkins1992transfer}, we introduce KC relation-based GNNs
to transfer the knowledge leveraging and changing throughout the three-stage learning process, as shown in Figure~\ref{fig:overview}.
To avoid repetition, we first elaborate on a prototype of KC relation-based GNNs in this section and highlight differences when applied to different stages in the subsequent sections.

Due to the lack of KC relation annotations, we follow previous works~\cite{nakagawa2019graph,song2022bi} that construct KC relations based on the data statistics.
Details could be referred to in Appendix~\ref{ap:graph_construction}.
Besides, we focus on the two most common relations, prerequisite and similarity
and extend three relation graphs ${\Pg,\Sg,\Rg}$, whose edges denote one KC being prerequisite/subsequent/relevant (similar) to another one.
This is because the forward and backward message passed along the unidirectional prerequisite relation should be differentiated.
Based on this, we design the KC relation-based GNNs with multiple layers.
They receive KC node features such as knowledge memory, knowledge gain/loss, or knowledge learnt in the three stages, which would be introduced later.
To capture the graph information, each layer first aggregates the features of each node's neighbors for each graph $\Gset$ from the last layer as 
\begin{equation}
\label{eq:gnn_layer_agg}
\bar{\textbf{f}}^{\Gg, (l)}_{c_i}=\frac{1}{|\mathcal{G}(c_i)|}\sum_{c_j\in\mathcal{G}(c_i)}\left(\beta^{\Gg}_{c_i,c_j}\cdot\tilde{\textbf{f}}^{(l-1)}_{c_j}\textbf{W}^{\mathcal{G}, (l)}_{proto}\right)
\end{equation} 
\begin{equation}
\label{eq:gnn_layer_fnn}
\tilde{\textbf{f}}^{\Gg,(l)}_{c_i} = \text{ReLU}\left(\bar{\textbf{f}}_{c_i}^{\Gg,(l)}\right)\textbf{O}_{proto}^{\mathcal{G},(l)}.
\end{equation}
where $\textbf{W}^{\mathcal{G}, (l)}_{proto}\in\real{d_{l-1}}{d_{l-1}}$ and $\textbf{O}_{proto}^{\mathcal{G},(l)}\in\real{d_{l-1}}{d_{l}}$ are the learnable weight matrices in this layer. $\Gg(\cdot)$ is the neighbor function of $\Gg$. $\text{ReLU}(\cdot)$ is an activation function to introduce non-linearity to enhance model representability.
$\beta^{\Gg}_{c_i,c_j}$ is the correlation score of KC $c_i$ and $c_j$ on graph $\Gg$, obtained by
\begin{equation}
\label{eq:question_question_score}
\beta^{\Gg}_{c_i,c_j}=\sigma\left(\trans{\textbf{k}}_{c_i}\textbf{W}^{\Gg}_{cor}\textbf{k}_{c_j}\right).
\end{equation}
$\textbf{k}_{c_i}, \textbf{k}_{c_j}\in\real{1}{d_e}$ are the two KCs' embeddings where $d_e$ is the number of embedding dimensions.
$\textbf{W}^{\Gg}_{cor}\in\real{d_e}{d_e}$ is the trainable matrix for $\Gg$, and $\sigma(\cdot)$ denotes the sigmoid function, which regularizes the score in $(0,1)$.
We then fuse the aggregated features from the three graphs by
\begin{equation}
\label{eq:gnn_layer_fuse}
\tilde{\textbf{f}}^{(l)}_{c_i} = 
\begin{cases} 
\sum_{\Gset}\tilde{\textbf{f}}_{c_i}^{\Gg,(l)} +  \tilde{\textbf{f}}^{(l-1)}_{c_i},& \text{if } d_{l-1}=d_l,\\
\sum_{\Gset}\tilde{\textbf{f}}_{c_i}^{\Gg,(l)},& \text{if } d_{l-1}\neq d_l,
\end{cases}
\end{equation}
where we apply a residual connection~\cite{szegedy2017inception} when $d_{l-1}=d_l$ to stabilize the training process.
In this prototype, we denote the input features of all KCs as $\tilde{\textbf{F}}^{(0)}\in\real{|C|}{d_0}$ and one of them as $\tilde{\textbf{f}}^{(0)}_{c_i}\in\real{1}{d_0}$ for KC $c_i$, and the output features as $\tilde{\textbf{F}}^{(L)}\in\real{|C|}{d_L}$ and $\tilde{\textbf{f}}_{c_i}^{(L)}\in\real{1}{d_L}$, where $d_0,d_L$ are the numbers of input and output feature dimensions.
Then this prototype GNN is formulated as:
\begin{equation}
    \label{eq:gnn_define_1}
\tilde{\textbf{F}}^{(L)}=\text{GNN}_{proto}(\tilde{\textbf{F}}^{(0)}|d_0,d_1,\cdots,d_L)
\end{equation}
\begin{equation}
    \label{eq:gnn_define_2}
    \tilde{\textbf{f}}_{c_i}^{(L)}=\text{GNN}_{proto}(\tilde{\textbf{f}}_{c_i}^{(0)}|d_0,d_1,\cdots,d_L).
\end{equation}
This prototype is then extended for different student learning stages to construct reasonable knowledge tracing based on the transfer of learning theory.
Besides, the number of layers $L$ controls the number of hops the feature propagates on the graphs, which clarifies the boundary between related and non-related KCs.

\subsection{Knowledge Memory \& Knowledge Tracing}

GRKT aims to model the process of student retrieving and learning knowledge with their memory.
Therefore, we employ a dynamic knowledge memory bank denoted as $\textbf{H}\in\real{|C|}{d_k}$, where each row $\textbf{h}_{c_i}$ encodes the current knowledge memory of KC $c_i$ for the student.
Here, $d_k$ signifies the number of memory dimensions.
This memory bank evolves alongside the student's learning process, represented as $\textbf{H}_t$, with a learnable initial state $\textbf{H}_0$ representing their prior knowledge before engaging in any learning behavior.
To track the knowledge mastery of a specific KC, we apply a non-negative projection vector $\textbf{w}_h\in\real{d_k}{1}_{\geq0}$ to $\textbf{h}_{c_i,t}$ using the equation:
\begin{equation}
\label{eq:mastery_projection}
\hat{m}_{c_i,t}=\textbf{h}_{c_i,t}\cdot\textbf{w}_h,
\end{equation}
which yields the mastery of KC $c_i$ at time step $t$.
The non-negative constraint on the network weights guarantees the monotonic relationship between mastery and each memory dimension. This technique has been widely adopted in numerous studies~\cite{wang2022neuralcd,wang2021using} to satisfy the monotonicity assumption.
Moreover, we leverage this constraint to establish a foundation for reasonable knowledge tracing, which would be gradually refined in subsequent descriptions.

\subsection{Stage I: Knowledge Retrieval}\label{subsec:kr}

In this stage, students retrieve stored knowledge from memory to solve given questions, a mechanism explained by memory theory~\cite{melton1963implications}.
Additionally, the transfer of learning theory~\cite{perkins1992transfer} suggests that learners transfer knowledge from similar fields to tackle problems. Leveraging this insight, we employ a KC relation-based GNN to model knowledge transfer from related KCs.
Specifically, we aggregate the knowledge memory of the given KC $c_t$ to solve its corresponding question $q_t$ before time step $t$ (represented as $t^-$):

\begin{equation}
\label{eq:knowledge_retrieval}
\tilde{\textbf{h}}_{c_t,t^-}^{(L)}=\text{GNN}_{rtv}(\tilde{\textbf{h}}^{(0)}_{c_t,t^-}|\{d_k\}_{L+1})\,,
\end{equation}
with initializing $\tilde{\textbf{h}}^{(0)}_{c_t,t^-} = \textbf{h}_{c_t,t^-}$.
Recognizing that different questions have different mastery requirements of KCs, we incorporate question-KC correlation scores into the aggregation process in this GNN, which are calculated by:
\begin{equation}
\label{eq:question_KC_scores}
\alpha_{q_i,c_j}=\sigma\left(\trans{\textbf{e}}_{q_i}\textbf{W}_{req}\textbf{k}_{c_j}\right)\,,
\end{equation}
where $\textbf{e}_{q_i}\in\real{d_e}{1}$ and $\textbf{k}_{c_j}$ are the embeddings of $q_i$ and $c_j$, and $\textbf{W}_{req}\in\real{d_e}{d_e}$ is a learnable matrix.
Then, the graph message aggregation process of Equation~\ref{eq:knowledge_retrieval} is actually
\begin{equation}
\label{eq:knowledge_retrieval_mod}
\tilde{\textbf{h}}^{\Gg, (l)}_{c_t}=\frac{1}{|\mathcal{G}(c_t)|}\sum_{c_i\in\mathcal{G}(c_t)}\left(\alpha_{q_t,c_i}\cdot\beta^{\Gg}_{c_t,c_i}\cdot\tilde{\textbf{h}}^{(l-1)}_{c_i}\textbf{W}^{\mathcal{G}, (l)}_{rtv}\right).
\end{equation}
We also remove the non-linear feed-forward process and restrict $\textbf{W}^{\mathcal{G}, (l)}_{rtv}\in\real{d_k}{d_k}_{\geq0}$ to ensure higher values of the related KCs' memory bring higher knowledge mastery.
After getting the aggregating knowledge memory from this GNN, we get the knowledge mastery as Equation~\ref{eq:mastery_projection} and compare it with the question difficulty $d_{q_t}$ to generate the predictive probability of solving the question:
\begin{equation}
\label{eq:predict}
\hat{a}_t = \sigma\left(\tilde{\textbf{h}}_{c_t,t}^{(L)}\cdot\textbf{w}_h-d_{q_t}\right).
\end{equation}
For multi-KC questions, we average the KCs' memory.  
The difficulty $d_{q_t}$ of question $q_t$ is generated by a Multi-Layer Perception (MLP):
\begin{equation}
\label{eq:difficulty}
d_{q_t}=\text{ReLU}\left(\bar{\textbf{e}}_{q_t}\textbf{W}^{(1)}_{diff}+\textbf{b}^{(1)}_{diff}\right)\textbf{W}^{(2)}_{diff}+\textbf{b}^{(2)}_{diff}.
\end{equation}
Here, $\bar{\textbf{e}}_{q_t}=[\textbf{k}_{c_t}\oplus\textbf{e}_{q_t}]$ is the concatenated representation of $q_t$ and its examined KC $c_t$'s embeddings.
For multi-KC questions, we use the KCs' average embedding.
$\textbf{W}_{diff}^{(1)}\in\mathbb{R}^{2d_e\times d_h}$, $\textbf{W}_{diff}^{(2)}\in\mathbb{R}^{d_h\times1}$, $\textbf{b}_{diff}^{(1)}\in\mathbb{R}^{1\times d_h}$, and $\textbf{b}_{diff}^{(2)}\in\mathbb{R}^{1\times 1}$ are learnable matrices and vectors.
$d_h$ is the number of hidden dimensions.
We denote the process of this two-layer MLP as $d_{q_t}=\text{MLP}_{diff}(\bar{\textbf{e}}_{q_t}|2d_e,d_h,1)$, and a similar notation is applied for brevity in subsequent descriptions.
Hereinafter, we accurately model the process whereby students retrieve knowledge from memory to answer new questions.

\subsection{Stage II: Memory Strengthening}
The testing effect theory~\cite{roediger2006test} reveals that a correct retrieval strengthens the storage of knowledge in memory, while an unsuccessful retrieval can lead to incorrect strengthening. Without correction or active learning after the error, this may reduce knowledge mastery~\cite{kornell2009unsuccessful}. 
In this stage, we determine the memory strengthening process based on whether the examined KC is correctly retrieved to solve the question, resulting in either knowledge gain or loss.
Additionally, these knowledge changes are propagated to related KCs based on the transfer of learning theory.
To enhance memory from a correct response to question $q_t$, we first combine and input the current memory $\textbf{h}_{c_t,t^-}$ of KC $c_t$ and the question information $\bar{\textbf{e}}_{q_t}$ into an MLP to obtain an initial memory feature:
\begin{equation}
\label{eq:memory_strengthening_mlp_gain}
\textbf{g}_{c_t,t}=\text{MLP}_{gain}\left([\textbf{h}_{c_t,t^-}\oplus\bar{\textbf{e}}_{q_t}]|d_k+2d_e,d_h,d_k\right).
\end{equation}
For multi-KC question, we calculate all the associated KCs' features.
This feature serves as a spark to propagate knowledge changes via another KC relation-based GNN.
Specifically, by initializing an input feature matrix $\tilde{\textbf{G}}^{(0)}_{t}$, where $\tilde{\textbf{g}}^{(0)}_{c_i,t}=\textbf{g}_{c_t,t}$ if $c_i=c_t$ and $\tilde{\textbf{g}}^{(0)}_{c_i,t} = \textbf{0}$ if $c_i\neq c_t$, the knowledge gain for all KCs is obtained as follows:
\begin{equation}
\label{eq:memory_strengthening_gnn_gain}
\tilde{\textbf{G}}_{t}^{(L)}=\text{ReLU}(\text{GNN}_{gain}(\tilde{\textbf{G}}^{(0)}_{t}|\{d_k\}_{L+1})).
\end{equation}

The $\text{ReLU}(\cdot)$ activation function ensures that the knowledge gain to be positive.
Moreover, due to the zero feature initialization except for the examined KC, the knowledge gain is only propagated to KCs within $L$ hops, delineating a boundary between related and unrelated KCs.
Similarly, we could derive the negative knowledge loss $\tilde{\textbf{L}}_{t}^{(L)}$ when students provide incorrect responses and wrongly strengthen their memory, by using a similar network $\text{GNN}_{loss}(\cdot)$.


Subsequently, we update the knowledge memory bank with respect to the response $a_t$ as follows:
\begin{equation}
\label{eq:memory_strengthening_update}
\textbf{H}_{t} = \textbf{H}_{t^-} + a_t\tilde{\textbf{G}}_{t}^{(L)} + (1-a_t)\tilde{\textbf{L}}_{t}^{(L)}.
\end{equation}
It is worth noting that different questions also have different effects on strengthening students' memory of KCs.
Therefore, similar to Equation~\ref{eq:knowledge_retrieval_mod}, these two GNNs also add the question-KC correlation scores during message passing.
Henceforth, the second stage, memory strengthening, is reasonably modeled based on the testing effect and the transfer of learning.

\subsection{Stage III: Knowledge Learning/Forgetting}
After students answer questions, their subsequent actions vary depending on the feedback received.
They may review their correct answers or correct their mistakes. 
Besides, they might prepare for the next question's KC they would encounter.
These active learning behaviors contribute to improving their knowledge mastery, which we model as the knowledge learning process in this stage.
Concretely, the KC of the last question and the next question both influence the student's learning target.
Therefore, we use an MLP to determine if the student actively learns them based on his/her current knowledge memory and the involved questions' information.
For KC $c_i\in\{c_t,c_{t+1}\}$ (or more involved KCs for multi-KC questions), the two-dimension policy distribution is calculated by:
\begin{equation}
\label{eq:knowledge_learning_decider}
\pi_{c_i,t}= \text{softmax}\left(\text{MLP}_{dcs}([\textbf{h}_{c_i,t}\oplus\bar{\textbf{e}}_{q_t}\oplus\bar{\textbf{e}}_{q_{t+1}}]|d_k+4d_e,d_h,2)\right).
\end{equation}
Here, $\text{argmax }\pi_{c_i,t}=0$ indicates that the first dimension is bigger.
We suppose there is no active learning.
Contrarily, $\text{argmax }\pi_{c_i,t}=1$ indicates the student would learn $c_i$.
Under this circumstance, we calculate the progress of learning $c_i$ in a similar way:
\begin{equation}
\label{eq:knowledge_learning_progress}
\textbf{p}_{c_i,t}=\text{MLP}_{prg}([\textbf{h}_{c_i,t}\oplus\bar{\textbf{e}}_{q_t}\oplus\bar{\textbf{e}}_{q_{t+1}}]|d_k+4d_e,d_h,d_k)).
\end{equation}
Based on the transfer of learning theory, this progress is also propagated to related KCs using another KC relation-based GNN. 
After initializing $\tilde{\textbf{P}}^{(0)}_{t}$ where $\tilde{\textbf{p}}^{(0)}_{c_i,t} = \textbf{p}_{c_i,t}$ for $c_i\in\{c_t,c_{t+1}\}$ with $\text{argmax }\pi_{c_i,t} = 1$, and $\tilde{\textbf{p}}^{(0)}_{c_i,t}=\textbf{0}$ otherwise, we compute
\begin{equation}
\label{eq:knowledge_learn_gnn}
\tilde{\textbf{P}}_{t}^{(L)}=\text{ReLU}(\text{GNN}_{prg}(\tilde{\textbf{P}}^{(0)}_{t}|\{d_k\}_{L+1})).
\end{equation}
This active learning process continues until the student answers the next question, allowing us to model each KC's progress $\tilde{\textbf{p}}_{c_i,t}^{(L)},c_i\in\Cset$ with a KC-specific time-aware kernel function to update:
\begin{equation}
    \label{eq:learning_progress_change}
    \textbf{h}_{c_i,(t+1)^-} = \textbf{h}_{c_i,t} + \boldsymbol{\phi}_{c_i}(\tilde{\textbf{p}}^{(L)}_{c_i,t},\Delta T_{t+1})
\end{equation}
where $\Delta T_{t+1} = T_{t+1}-T_{t}$ is the time duration until the next question.
According to the learning curve~\cite{yelle1979learning}, the efficiency of students in learning a specific KC tends to be high initially and gradually decreases over both the learning time and frequency.
Therefore, we design the kernel function in an exponential form:
 \begin{equation}
        \label{eq:learning_progress_kernel}
    \boldsymbol{\phi}_{c_i}(\tilde{\textbf{p}}^{(L)}_{c_i,t},\Delta T_{t+1})=\tilde{\textbf{p}}^{(L)}_{c_i,t}\odot(\textbf{1}-\text{exp}(-(n_{c_i,t}+1)\Delta T_{t+1}\cdot\tilde{\boldsymbol{\gamma}}_{c_i}^{(L)})),
    \end{equation}
where $\odot$ is the Hadamard product. $n_{c_i,t}$ is the number of times that $c_i$ has been learned by the student.
$\tilde{\boldsymbol{\gamma}}_{c_i}^{(L)}$ represents the KC-specific kernel parameters of $c_i$ generated by 
another KC relation-based GNN.
It leverages the property of graph homophily that makes related KCs have similar learning ratios:

\begin{equation}
\label{eq:knowledge_progress_gnn}
\tilde{\boldsymbol{\gamma}}_{c_i}^{(L)}=\text{softplus}(\text{GNN}_{lrn}(\tilde{\boldsymbol{\gamma}}_{c_i}^{(0)}|d_e,\{d_k\}_{L}))\,,
\end{equation}
with initializing $\tilde{\boldsymbol{\gamma}}_{c_i}^{(0)} = \textbf{k}_{c_i}$ which is $c_i$'s embedding.
Here, $\text{softplus}(\cdot)$ is an activation function to restrict the parameter to be positive.
On the other hand, for KCs that students have acquired before but they do not choose to learn, we introduce the knowledge forgetting process.
Therefore, for the KCs students do not make progress on (i.e., $\tilde{\textbf{p}}^{(L)}_{c_i,t}=\textbf{0}$), their previously acquired knowledge fades over time:
\begin{equation}
    \label{eq:forgetting_progress_change}
    \textbf{h}_{c_i,(t+1)^-} = \textbf{h}_{c_i,t} - \boldsymbol{\kappa}_{c_i}(\Delta\textbf{h}_{c_i,t},\Delta T_{t+1})
\end{equation}
where $\Delta\textbf{h}_{c_i,t} = \textbf{h}_{c_i,t} - \textbf{h}_{c_i,0}$ represents the total knowledge acquisition the student has accumulated.
According to the forgetting curve~\cite{ebbinghaus1885gedachtnis}, the speed that students forget knowledge follows a pattern of initially rapid decay and then a gradual decrease over time and the review frequency.
Therefore, we similarly design KC-specific forgetting kernel functions in an exponential form:
\begin{equation}
\label{eq:learning_forgetting_kernel}
\boldsymbol{\kappa}_{c_i}(\Delta\textbf{h}_{c_i,t},\Delta T_{t+1})=\Delta\textbf{h}_{c_i,t}\odot(\textbf{1}-\text{exp}(-(n_{c_i,t}+1)\Delta T_{t+1}\cdot\tilde{\boldsymbol{\theta}}_{c_i}^{(L)})),
\end{equation}
where the kernel parameters $\tilde{\boldsymbol{\theta}}_{c_i}^{(L)}$ are similarly generated by another KC relation-based GNN:
\begin{equation}
\label{eq:knowledge_forgetting_gnn}
\tilde{\boldsymbol{\theta}}_{c_i}^{(L)}=\text{softplus}(\text{GNN}_{fgt}(\tilde{\boldsymbol{\theta}}_{c_i}^{(0)}|d_e,\{d_k\}_{L}))
\end{equation}
with initializing $\tilde{\boldsymbol{\theta}}_{c_i}^{(0)} = \textbf{k}_{c_i}$.
Consequently, based on the learning and forgetting curves, we have derived the updated knowledge memory $\textbf{H}_{(t+1)^-}$ in this stage, which is recursively used for answering the next question.

\begin{table}[!t]
\caption{Statistics of the three preprocessed datasets.}
\label{tab:stat}
\begin{tabular}{l|ccc}
\hline
Dataset            & ASSIST09 & ASSIST12 & Junyi  \\ \hline
\#response         & 0.4m     & 2.6m     & 25.4m  \\
\#sequence         & 7.4k    & 38.1k    & 325.4k \\
\#question         & 13.5k    & 51.0k    & 2.8k  \\
\#concept          & 140      & 198      & 722    \\
\#concept/question & 1.22     & 1.0        & 1.0    \\ 
\hline
\end{tabular}
\end{table}


\begin{table*}[!t]
\setlength{\tabcolsep}{1.2pt}

\caption{Results of the main experiments. The best results among GRKT and the baselines are in bold. The second ones are in italic. * indicates statistical significance over the best baseline, measured by T-test with p-value $\leq$ 0.05. ``CONS'', ``GAUC'' and ``RPT'' are short for the three metrics for reasonability, consistency, GAUCM and Repetition.}

\begin{tabular}{l|ccccc|ccccc|ccccc}
\hline
Dataset & \multicolumn{5}{c|}{ASSIST09}                                                                                               & \multicolumn{5}{c|}{ASSIST12}                                                                                              & \multicolumn{5}{c}{Junyi}                                                                                                         \\ \hline
Metric  & \multicolumn{1}{c}{AUC} & \multicolumn{1}{c}{ACC} & \multicolumn{1}{c}{CONS} & \multicolumn{1}{c}{GAUC} & \multicolumn{1}{c|}{RPT} & \multicolumn{1}{c}{AUC} & \multicolumn{1}{c}{ACC} & \multicolumn{1}{c}{CONS} & \multicolumn{1}{c}{GAUC} & \multicolumn{1}{c|}{RPT} & \multicolumn{1}{c}{AUC} & \multicolumn{1}{c}{ACC} & \multicolumn{1}{c}{CONS} & \multicolumn{1}{c}{GAUC} & \multicolumn{1}{c}{RPT} \\ \hline
DKT     & 0.7695                  & 0.7246                  & 0.6463                   & 0.7172                   & 0.8131                   & 0.7303                  & 0.7358                  & 0.6772                   & 0.6929                   & 0.7955                   & 0.8003                  & 0.8541                  & 0.7432                   & 0.6415                   & 0.8790                  \\
DKVMN    & 0.7680                  & 0.7239                  & 0.8708                   & 0.7116                   & 0.8061                   & 0.7279                  & 0.7349                  & 0.9273                   & 0.6729                   & 0.7971                   & 0.8004                  & 0.8541                  & 0.9455                   & 0.6379                  & 0.8780                 \\
DKT+    & 0.7707                  & 0.7245                  & 0.6364                   & 0.7089                   & \underline{0.8395}                   & 0.7300                  & 0.7353                  & 0.6809                   & 0.6766                   & 0.8172                   & 0.7993                  & 0.8539                  & 0.7624                   & \underline{0.6436}                   & 0.8869                  \\
SAKT    & 0.7634                  & 0.7206                  & 0.8539                   & 0.7101                   & 0.7749                   & 0.7227                  & 0.7329                  & 0.8202                   & 0.6866                   & 0.7797                   & 0.7995                  & 0.8535                  & 0.8600                   & 0.6387                   & 0.8747                  \\
GKT     & 0.7702                  & 0.7252                  & 0.6697                   & \underline{0.7183}                   & 0.8124                   & 0.7339                  & 0.7372                  & 0.7450                   & 0.6971                   & 0.7986                   & 0.8023                  & 0.8547                  & 0.7403                   & 0.6398                   & 0.8788                  \\
AKT     & 0.7820                  & 0.7320                  & 0.5870                   & 0.7113                   & 0.8184                   & 0.7665                  & 0.7514                  & 0.5909                   & 0.6892                   & 0.8172                   & 0.8161                  & 0.8593                  & 0.5810                   & 0.6398                   & 0.8734                  \\
SKT     & 0.7732 & 0.7273 & 0.7023& 0.7098& 0.8092                   & 0.7354 & 0.7398 & 0.7813 & 0.6952 & 0.7934                   & 0.8045 &0.8552 & 0.7792 & 0.6420 & 0.8805                  \\
LPKT    & \underline{0.7869}                  & 0.7369                  & 0.7909                   & 0.7124                   & 0.8205                   & 0.7740                  & 0.7556                 & 0.8174                   & 0.6839                   & \underline{0.8255}                   & 0.8153                  & 0.8585                  & 0.7238                   & 0.6453                   & 0.8845                  \\
DIMKT   & 0.7814                  & 0.7351                  & 0.7899                   & 0.7153                   & 0.8221                   & 0.7711                  & 0.7550                  & 0.8099                   & \underline{0.6995}                   & 0.8198                   & \underline{0.8163}                  & \underline{0.8594}                  & 0.8945                   & 0.6424                   & 0.8850                  \\

DTrans  & 0.7858                  & 0.7345                  & \underline{0.8928}                   & 0.7126                   & 0.8253                   & 0.7720                  & 0.7542                  & \underline{0.9217}                   & 0.6863                   & 0.8249                   & 0.8149                  & 0.8577                  & \underline{0.9274}                   & 0.6420                   & \underline{0.8893}                  \\ LBKT  & 0.7865 & \underline{0.7372} & 0.8054 & 0.7134 &0.8225                   & \underline{0.7763} & \underline{0.7562} & 0.8123 & 0.6814 &0.8230                   & 0.8140 & 0.8568 & 0.8123 & 0.6409 & 0.8871   \\ \hline
GRKT     & \textbf{0.7914*}                  & \textbf{0.7398*}                  & \textbf{1.0000*}                   & \textbf{0.7209*}                   & \textbf{0.8486*}                   & \textbf{0.7794*}                  & \textbf{0.7576}                  & \textbf{1.0000*}                   & \textbf{0.7064*}                   & \textbf{0.8319*}                   & \textbf{0.8207*}                  & \textbf{0.8624*}                  & \textbf{1.0000*}                   & \textbf{0.6473*}                   & \textbf{0.8957*}                  \\ \hline
improv. & 0.57\%                  & 0.35\%                  & 12.01\%                  & 0.36\%                   & 1.09\%                   & 0.40\%                  & 0.19\%                  & 8.50\%                   & 0.98\%                   & 0.78\%                   & 0.54\%                  & 0.35\%                  & 7.83\%                   & 0.31\%                   & 0.72\%                  \\ \hline
\end{tabular}
\label{tab:main_exp}
\end{table*}

\begin{table*}[!t]
\setlength{\tabcolsep}{2.5pt}
\caption{Results of the ablation experiments.}

\begin{tabular}{l|ccccc|ccccc|ccccc}
\hline
Dataset & \multicolumn{5}{c|}{ASSIST09}                                                                                               & \multicolumn{5}{c|}{ASSIST12}                                                                                              & \multicolumn{5}{c}{Junyi}                                                                                                         \\ \hline
Metric  & \multicolumn{1}{c}{AUC} & \multicolumn{1}{c}{ACC} & \multicolumn{1}{c}{CONS} & \multicolumn{1}{c}{GAUC} & \multicolumn{1}{c|}{RPT} & \multicolumn{1}{c}{AUC} & \multicolumn{1}{c}{ACC} & \multicolumn{1}{c}{CONS} & \multicolumn{1}{c}{GAUC} & \multicolumn{1}{c|}{RPT} & \multicolumn{1}{c}{AUC} & \multicolumn{1}{c}{ACC} & \multicolumn{1}{c}{CONS} & \multicolumn{1}{c}{GAUC} & \multicolumn{1}{c}{RPT} \\ \hline
GRKT     & \textbf{0.7914}                  & \textbf{0.7398}                  & \textbf{1.0000}                   & \textbf{0.7209}                   & \textbf{0.8486}                   & \textbf{0.7794}                 & \textbf{0.7576}                  & \textbf{1.0000}                   & \textbf{0.7064}                   & \textbf{0.8319}                   & \textbf{0.8207}                  & \textbf{0.8624}                  & \textbf{1.0000}                   & \textbf{0.6473}                   & \textbf{0.8957}                  \\ \hline
-LF  & 0.7871                  & 0.7367                  & 1.0000                   & 0.7066                   & 0.8243                   & 0.7767                  & 0.7558                  & 1.0000                   & 0.6809                   & 0.8276                   & 0.8170                  & 0.8598                  & 1.0000                   & 0.6401                   & 0.8815                  \\
-SIM-PRE & 0.7578                  & 0.7161                  & 1.0000                   & 0.6197                   & 0.8246                   & 0.7502                  & 0.7424                  & 1.0000                   & 0.6223                   & 0.8291                   & 0.7921        & 0.8481        & 1.0000         & 0.6084         & 0.8781        \\
-SIM & 0.7896                  & 0.7375                  & 1.0000                   & 0.7135                   & 0.8402                   & 0.7777                  & 0.7564                  & 1.0000                   & 0.6888                   & 0.8259                   & 0.8186                  & 0.8611                  & 1.0000                   & 0.6447                   & 0.8862                  \\
-PRE & 0.7897                  & 0.7384                  & 1.0000                   & 0.7149                   & 0.8437                   & 0.7779                  & 0.7563                  & 1.0000                   & 0.6915                   & 0.8264                   & 0.8191                  & 0.8615                  & 1.0000                   & 0.6452                   & 0.8897                  \\ \hline
\end{tabular}
\label{tab:abl_exp}
\vspace{-0.8em}
\end{table*}

\subsection{Model Training}
The three-stage modeling is recurrent along the student response sequence.
After learning/forgetting knowledge in the third stage, the updated knowledge memory is prepared for the first stage to answer the next question.
This makes GRKT an end-to-end style so we directly train the model by the binary cross-entropy loss, aligning the predictive probability $\hat{a}^u_t$ from Equation~\ref{eq:predict} with the ground-truth response correctness label $a^u_t$:
\begin{equation}
\label{eq:bce_loss}
\mathcal{L}=-\sum_{u\in\Uset}\sum_{r_t^u\in\Hset^u}a^u_t\log\hat{a}^u_t+(1-a^u_t)\log(1-\hat{a}^u_t).
\end{equation}
Here, we omit the averaging notation for brevity.
Besides, we also apply the $l_2$ normalization to the model parameters during the training process to avoid the over-fitting issue.

\section{Experiments}
In this section, we design comprehensive experiments to address the following research questions:
\begin{itemize}[leftmargin=0.7cm]
\item[\textbf{Q1:}] Does GRKT achieve competitive results in terms of both prediction performance and knowledge tracing reasonability compared to current state-of-the-art DLKT methods?
\item[\textbf{Q2:}] What are the roles and impacts of different components of GRKT on the overall performance and reasonability?
\item[\textbf{Q3:}] How reasonable is the knowledge mastery traced by GRKT from an intuitive perspective?
\end{itemize}
Additionally, we conduct other experiments such as hyper-parameter analysis. Due to space constraints, we include them in Appendix~\ref{ap:hyper_analysis}.

\subsection{Experimental Setup}
\subsubsection{Datasets}\label{sec:dataset} We evaluate the performance of GRKT on three widely-used public KT datasets:
\begin{itemize}[leftmargin=*]
\item \textbf{ASSIST09}~\cite{feng2009addressing}\footnote{\url{https://sites.google.com/site/assistmentsdata/home/2009-2010-assistment-data}}: ASSISTments is an online tutoring system for mathematics, which collected this dataset from 2009 to 2010. We use the \textit{combined} version. For the missing timestamp information, we approximate it using the field \textit{order\_id}.
\item \textbf{ASSIST12}~\cite{feng2009addressing}\footnote{\url{https://sites.google.com/site/assistmentsdata/home/2012-13-school-data-with-affect}}: Another dataset from ASSISTments, collected during the period of 2012 to 2013.
\item \textbf{Junyi}~\cite{chang2015modeling}\footnote{\url{https://pslcdatashop.web.cmu.edu/Files?datasetId=1198}}: This dataset is collected from the Junyi Academy online platform in 2015. It contains a part of annotated KC relationships which are suitable for the requirements of GRKT. We use the \textit{junyi\_ProblemLog\_original.csv} version.
\end{itemize}

For preprocessing each dataset, we partition the response sequences of every student into subsequences, each containing 100 responses.
Subsequences containing fewer than 10 responses are eliminated, while those with less than 100 responses are padded with zeros to meet the required length. 
Statistics of the processed datasets can be found in Table~\ref{tab:stat}.

\subsubsection{Evaluation}
As a binary classification task of predicting student responses, we utilize the area under the curve (AUC) and accuracy (ACC) as the evaluation metrics for prediction performance.
For evaluating model reasonability, we introduce three metrics:

\begin{itemize}[leftmargin=*]
    \item \textbf{Consistency}: We propose this metric to measure the ratio of consistent variation between the mastery of KCs. When a student's mastery of the corresponding KC declines after answering a certain question, the mastery of other KCs should either decline (for related KCs) or remain unchanged (for unrelated KCs). We calculate this percentage.
    \item \textbf{GAUCM}: This metric calculates the average AUC scores with respect to the mastery of each question's examined KC. Its reflects the monotonicity assumption: a question could be more likely to be correctly answered if students have higher mastery of its KC. This metric is proposed by Zhang et al.~\cite{zhang2023counterfactual}.
    \item \textbf{Repetition}: This metric is proposed by Yeung et al.~\cite{yeung2018addressing}, stating that a reasonable KT method should satisfy: after a student has finished a question and is given this same question again, the response result (correct or incorrect) should remain the same. We calculate the accuracy under this circumstance.
\end{itemize}
The formulas of these metrics are presented in Appendix~\ref{ap:reasonability_formula}.
Moreover, we employ a five-fold cross-validation to assess the model's performance.
10\% of the sequences of each fold serve as the validation set for parameter tuning.
We stop the training when the validation performance fails to improve for 10 consecutive epochs.

\subsubsection{Baselines}
To compare with mainstream DLKT methods covering different aspects, we select eleven baselines from 2015 to 2023, including DKT~\cite{piech2015deep}, DKVMN~\cite{zhang2017dynamic}, DKT+~\cite{yeung2018addressing}, SAKT~\cite{pandey2019self}, GKT~\cite{nakagawa2019graph}, AKT~\cite{ghosh2020context}, SKT~\cite{tong2020structure}, LPKT~\cite{shen2021learning}, DIMKT~\cite{shen2022assessing}, Dtransformer~\cite{yin2023tracing} and LBKT~\cite{xu2023learning}.
Among them, GKT and SKT leverages the KC graph, and LPKT leverage the timestamp information.
DKT+, LPKT and Dtransformer consider some aspects of model reasonability: the knowledge tracing stability or learning/forgetting behaviors, but not comprehensively address the DLKT unreasonableness issue.
For the methods not providing the proxy of tracing knowledge mastery, AKT and DIMKT, we follow previous works~\cite{cui2023fine,liu2019ekt} that replace input question features with zeros to estimate the mastery.
We note that cognitive diagnosis baselines are not considered because they usually focus on static testing environments~\cite{leighton2007cognitive} but we study in the
dynamic learning situation.
\subsubsection{Implementation Details}
We employ the Adam optimizer~\cite{kingma2014adam} for all methods to achieve their best performance. We choose their learning rates from \{1e-2, 5e-3, 1e-3, 5e-4, 1e-4\}, and fixed the embedding and hidden dimension numbers at 128 for fairness.
We strictly follow the original papers of all methods to set their hyper-parameters.
For GRKT, detailed hyper-parameter setting is referred in the Appendix~\ref{ap:hyper_setting}.
Furthermore, for the non-negative constraint on the specified network weights in Equations~\ref{eq:mastery_projection} and \ref{eq:knowledge_retrieval_mod}, we use the softmax operation along the knowledge memory dimension, which performs best in practice.
Besides, the Junyi dataset includes some labeled relations, which we experiment with and present the results in Appendix~\ref{ap:junyi}.

\subsection{Overall Performance (Q1)}

Table~\ref{tab:main_exp} illustrates the comprehensive performance comparison between GRKT and eleven other baselines.
Notably, GRKT showcases the highest efficacy, surpassing the leading baselines by margins ranging from 0.19\% to 12.01\% across both prediction performance and reasonability metrics.
For metrics such as AUC and ACC, which primarily gauge predictive accuracy, the state-of-the-art DLKT techniques, LPKT, and DIMKT exhibit exemplary performance owing to their sophisticated neural architectures.
Besides, methods that emphasize aspects of reasonability, such as enhancing knowledge tracing stability and explicitly modeling learning and forgetting behaviors, DKT+, LPKT, and DTransformer, demonstrate competitive performance across reasonableness metrics.
These methods secure seven out of nine second-place positions in reasonability metrics.
Remarkably, GRKT achieves a perfect score of 1.0 on the consistency metric, signifying its ability to effectively address the challenge of maintaining consistency in knowledge mastery changes across KCs by the network constraints.

\begin{figure*}[!t]
  \centering
  \includegraphics[width=1.0\linewidth]{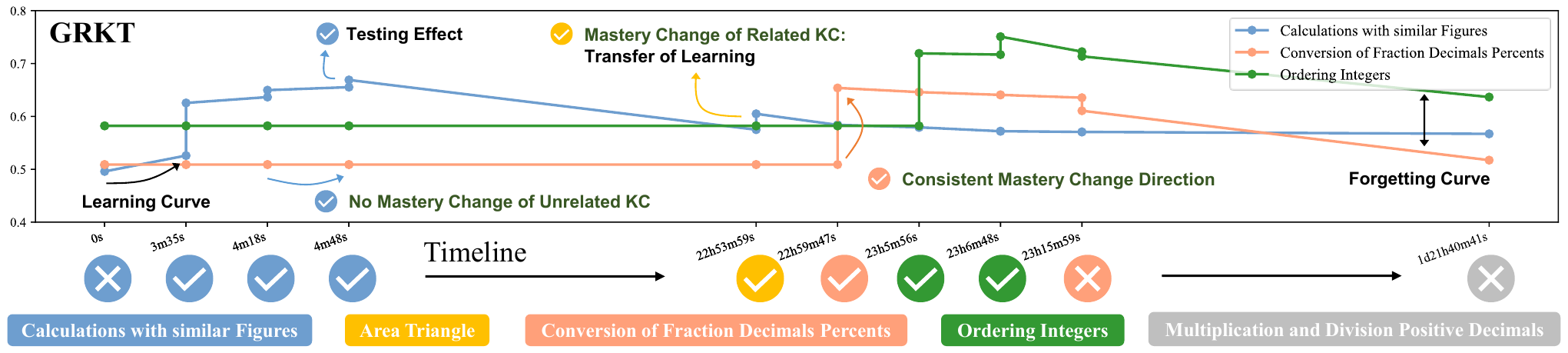}
  \caption{Case study of the same student's evolving knowledge mastery exemplified in Section~\ref{sec:intro}.}
 \label{fig:case}
\end{figure*}

\begin{figure*}[!t]
  \centering
  \includegraphics[width=1.0\linewidth]{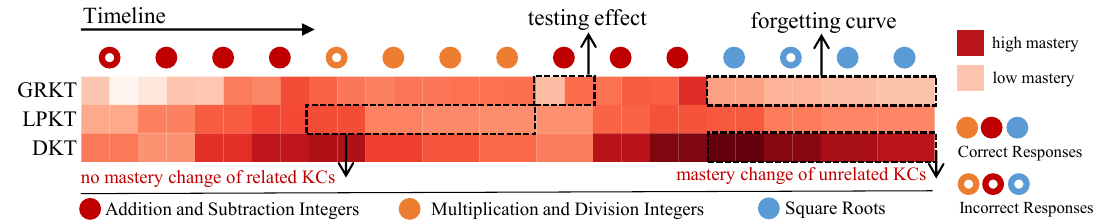}
  \caption{Knowledge tracing heatmap of GRKT, LPKT and DKT tracing one another student’s mastery on KC \textit{Addition and Subtraction Integer}s. Different colors represent different KCs.}
 \label{fig:heatmap}
\end{figure*}

\subsection{Ablation Study (Q2)}
The ablation study aims to evaluate the impact of each component in GRKT by removing specific techniques and comparing the results with the full model. Four components are removed:
\begin{itemize}[leftmargin=*]
    \item \textbf{-LF}: Removal of the third stage, knowledge learning/forgetting.
    \item \textbf{-SIM}: Removal of the similarity relation.
    \item \textbf{-PRE}: Removal of the prerequisite relation.
    \item \textbf{-SIM-PRE}: Removal of the leverage of KC relation graphs.
\end{itemize}

As shown in Table~\ref{tab:abl_exp}, GRKT-SIM-PRE experiences the most significant deterioration, emphasizing the crucial role of KC relations in the KT task. Moreover, when only one of these two relations is utilized, there is a notable improvement in performance, indicating that each provides meaningful information for GRKT. 
Moreover, the performance is further enhanced when both relations are used together.
Additionally, the degradation of GRKT-LF underscores the importance of modeling the knowledge learning/forgetting stage.

\subsection{Reasonable Knowledge Tracing (Q3)}

To intuitively validate the resonability of GRKT, we present one student's dynamic knowledge mastery traced by GRKT in Figure~\ref{fig:case}.
As depicted, the result aligns well with our hypothesis of a comprehensive and reasonable knowledge tracing model integrating various effects based on pedagogical theories.
Furthermore, it addresses three key issues in the reasonableness of existing DLKT methods: mastery changes of unrelated KCs, not mastery changes of related KCs, and inconsistent mastery change direction.
We also present GRKT, LPKT and DKT tracing one another student's mastery on KC \textit{Addition and Subtraction Integers} in Figure~\ref{fig:heatmap}.
As shown, GRKT yields reasonable knowledge tracing results such as the fine-grained knowledge changing from testing effects and the faded
knowledge with forgetting curves. LPKT and DKT still have reasonable issues such as the mastery change of unrelated KCs and no mastery change of related KCs.

\subsection{Complexity Analysis}

Although the detailed methodology description of GRKT, its internal composition of only GNNs and MLPs does not make the inference complicated.
Suppose $t$ is the length of response sequence, $C$ is the KC set, $E$ is the KC relation edge set, $d$ is the hidden dimension number we set as a small value of 16, and $k$ is the GRKT's memory dimension number.
The time complexity of GRKT is then $O(t|E|k + |E|d + t|C|k^2 + |C|d^2 + td^2)$, consisting of feature aggregation $O(t|E|k + |E|d)$ and feature non-linear transformation $O(t|C|k^2 + |C|d^2)$ of the GNNs, and $O(td^2)$ of the MLPs.
In contrast, other comparable attention or RNN-based methods usually have time complexity $O(td^2 + t^2d)$.
In real scenarios, $t,|C|,d$ usually lie in 100-200 and the KC relation graphs are sparse.
Therefore, we can approximately assume $t=d=|C|=k^2=n$ and $|E| = k \cdot |C|$ to facilitate the complexity comparison, which indicates the GRKT's time complexity is actually in the same order of magnitude $O(n^3)$ as other methods.
We also test the inference speed of GRKT. It averagely costs 60ms for one student, which is acceptable in practice.

\section{conclusion}

In this paper, we point out the issue that many existing DLKT approaches prioritize predictive accuracy over tracking students' dynamic knowledge mastery. This often results in models that yield unreasonable outcomes, complicating their application in real teaching scenarios.
To this end, our study introduces GRKT, a graph-based reasonable knowledge tracing. It employs graph neural networks and consists of a finer-grained three-stage modeling process based on pedagogical theories, conducting a more reasonable knowledge tracing. Extensive experiments across multiple datasets demonstrate that GRKT not only enhances predictive accuracy but also generates more reasonable knowledge tracing results.
In the future, we plan to address certain limitations of GRKT, such as enhancing the model's ability to provide more fine-grained responses, including multiple-choice or essay answers.
Furthermore, we would evaluate GRKT in real teaching scenarios.

\bibliographystyle{ACM-Reference-Format}
\bibliography{sample-base}

\appendix

\begin{table}[!t]
\setlength{\tabcolsep}{1.2pt}

\caption{The notation table of GRKT. We omit the superscript of the target student $u$ whose knowledge is to be traced.}
\label{tab:notation}
\begin{tabular}{c|c}
\hline
 \multicolumn{2}{c}{Task formulation}\\
 \hline
$\Uset,\Qset,\Cset$ & sets of students, questions, KCs\\
$c_i,c_j,q_i,q_j$ & certain KCs, questions\\
$u, t$ & the target student, time step\\
$\Hset$ & response history of $u$\\
$r_t$ & response of $u$ at $t$\\
$q_t,c_t$ & question and examined KC of $r_t$\\
$a_t,T_t$ & binary correctness and timestamp of $r_t$\\
$\Mset$& evolving knowledge mastery of $u$\\
$\textbf{m}_{t}$& knowledge mastery of $u$ at $t$\\
$m_{c_i,t}$& knowledge mastery of $c_i$ of $u$ at $t$\\
\hline
\multicolumn{2}{c}{KC relation-based GNN}\\
\hline
$\Pg,\Sg,\Rg$ & prerequisite, subsequence, similarity graphs\\
$\Pg(\cdot),\Sg(\cdot),\Rg(\cdot)$ & neighbor functions of $\Pg,\Sg,\Rg$\\
$\Gg$ & certain graph in $\Pg,\Sg,\Rg$\\
$\Gg(\cdot)$ & neighbor function of $\Gg$\\
$L$& number of GNN layers\\
$\text{GNN}_{proto}$& prototype of KC relation-based GNN\\
$d_0,d_1,...,d_L$& \# of dimensions of prototype GNN's layers\\
$\tilde{\textbf{f}}_{c_i}^{(0)},\tilde{\textbf{F}}^{(0)}$ & prototype input of $c_i$ and all to $\text{GNN}_{proto}$\\
$\tilde{\textbf{f}}_{c_i}^{(L)},\tilde{\textbf{F}}^{(L)}$ & prototype output of $c_i$ and all from $\text{GNN}_{proto}$\\
\multirow{2}{*}{$\tilde{\textbf{f}}_{c_i}^{(l)},\tilde{\textbf{F}}^{(l)}$} & prototype intermedium of $c_i$ and all of \\
& $l^{th}$ layer of $\text{GNN}_{proto}$\\
$\textbf{W}^{\mathcal{G}, (l)}_{proto},\textbf{O}^{\mathcal{G}, (l)}_{proto}$ & weight matrices 
of $l^{th}$ of $\text{GNN}_{proto}$ for $\Gg$ \\

\multirow{2}{*}{$\tilde{\textbf{f}}_{c_i}^{\Gg,(l)}$} & prototype intermedium of $c_i$ of $l^{th}$ layer \\
& of $\text{GNN}_{proto}$ for $\Gg$\\\hline
\multicolumn{2}{c}{GRKT basic factors}\\\hline
$\textbf{e}_{q_i},\textbf{e}_{q_t}, \textbf{k}_{c_i}, \textbf{k}_{c_t}$ & embeddings of $q_i, q_t, c_i, c_t$\\
$\bar{\textbf{e}}_{q_i},\bar{\textbf{e}}_{q_t}$ & concatenation of $q_i$ and its KC's embeddings\\
$\alpha_{q_t, c_j}$ & requirement score of $q_t$ requiring $c_j$\\
$\textbf{W}_{req}$ & matrix to calculate requiring scores\\
$\beta^{\Gg}_{c_i, c_j}$ & correlation score of $c_i$ and $c_j$ for $\Gg$\\
$\textbf{W}^{\Gg}_{cor}$ & matrix to calculate correlation scores for $\Gg$\\
$\textbf{H}_{0}$ & initial knowledge memory of $u$\\
$\textbf{H}_{t^{-}}$ & knowledge memory of $u$ at a moment before $t$\\
$\textbf{H}_{t}$ & knowledge memory of $u$ at $t$\\
$\textbf{h}_{c_i,t}$ & knowledge memory of $c_i$ of $u$ at $t$\\
$d_e, d_k, d_h$ & \ embedding, memory, and hidden dimensions\\
$\textbf{w}_{h}$ & vector to project knowledge memory to mastery\\
$\hat{m}_{c_i,t}$ & modeled knowledge mastery of $c_i$ at $t$\\
$d_{q_t}$ & question difficulty of $q_t$\\
$\text{MLP}_{diff}$ & MLP to generate question difficulty\\
$\textbf{W}_{diff}^{(1)},\textbf{W}_{diff}^{(2)}$ & weight matrices in $\text{MLP}_{diff}$\\
$\textbf{b}_{diff}^{(1)},\textbf{b}_{diff}^{(2)}$ & weight vectors in $\text{MLP}_{diff}$\\
\hline
\end{tabular}
\vspace{-1em}
\end{table}

\begin{table}[!t]
\setlength{\tabcolsep}{1.2pt}
\caption{The continuing notation table for the three stages.}
\label{tab:notation2}
\begin{tabular}{c|c}
\hline
\multicolumn{2}{c}{Stage I: knowledge retrieval}\\\hline
$\text{GNN}_{rtv}$&  KC relation-based GNN for knowledge retrieval\\
$\tilde{\textbf{h}}_{c_i,t^-}^{(0)},\tilde{\textbf{H}}_{t^-}^{(0)}$ & memory input of $c_i$ and all to $\text{GNN}_{rtv}$ before $t$\\
$\tilde{\textbf{h}}_{c_i,t^-}^{(L)},\tilde{\textbf{H}}^{(L)}_{t^-}$ & memory output of $c_i$ and all from $\text{GNN}_{rtv}$ before $t$\\
$\hat{a}_t$ & predictive probability of $a_t$\\\hline
\multicolumn{2}{c}{Stage II: memory strengthening}\\\hline
$\text{MLP}_{gain}$ & MLP to get memory feature for knowledge gain\\
$\textbf{g}_{c_t,t}$ & memory feature of $c_t$ at $t$ for knowledge gain\\
$\text{GNN}_{gain}$&  KC relation-based GNN for knowledge gain\\
$\tilde{\textbf{g}}_{c_t,t}^{(0)},\tilde{\textbf{G}}_t^{(0)}$ & memory feature input of $c_t$ and all to $\text{GNN}_{gain}$ at $t$\\
$\tilde{\textbf{g}}_{c_t,t}^{(L)},\tilde{\textbf{G}}_t^{(L)}$ & knowledge gain of $c_t$ and all from $\text{GNN}_{gain}$ at $t$\\
$\text{MLP}_{loss}$ & MLP to get memory feature for knowledge loss\\
$\textbf{l}_{c_t,t}$ & memory feature of $c_t$ at $t$ for knowledge loss\\
$\text{GNN}_{loss}$&  KC relation-based GNN for knowledge loss\\
$\tilde{\textbf{l}}_{c_t,t}^{(0)},\tilde{\textbf{L}}_t^{(0)}$ & memory feature input of $c_t$ and all to $\text{GNN}_{loss}$ at $t$\\
$\tilde{\textbf{l}}_{c_i,t}^{(L)},\tilde{\textbf{L}}_t^{(L)}$ & knowledge loss of $c_t$ and all from $\text{GNN}_{loss}$ at $t$\\\hline
\multicolumn{2}{c}{Stage III: knowledge learning/forgetting}\\\hline
$\text{MLP}_{dsc}$ & MLP to get policy distribution for active learning\\
$\pi_{c_i,t}$ & policy distribution if $u$ decide to learn $c_i$ at $t$\\
$\text{MLP}_{prg}$ & MLP to get initial knowledge progress\\
$\textbf{p}_{c_i,t}$ & initial knowledge progress of $c_i$ at $t$\\
$\text{GNN}_{prg}$& KC relation-based GNN for knowledge progress\\
$\tilde{\textbf{p}}_{c_i,t}^{(0)},\tilde{\textbf{P}}_t^{(0)}$ & initial progress input of $c_i$ and all to $\text{GNN}_{prg}$ at $t$\\
$\tilde{\textbf{p}}_{c_i,t}^{(L)},\tilde{\textbf{P}}_t^{(L)}$ & knowledge progress of $c_i$ and all from $\text{GNN}_{prg}$ at $t$\\
$\Delta T_{t+1}$ & time interval between $T_{t}$ and $T_{t+1}$\\
$\boldsymbol{\phi}_{c_i}$ & KC-specific time-aware kernel for learning $c_i$\\
$n_{c_i, t}$ & \# of times $u$ has learnt $c_i$\\
$\text{GNN}_{lrn}$& KC relation-based GNN to get parameters of $\boldsymbol{\gamma}_{c_i}$\\
$\tilde{\boldsymbol{\gamma}}_{c_i}^{(0)}$ & input feature of $c_i$ initialized as $\textbf{k}_{c_i}$ to $\text{GNN}_{lrn}$\\
$\tilde{\boldsymbol{\gamma}}_{c_i}^{(L)}$ & output parameters of $\boldsymbol{\gamma}_{c_i}$ for $c_i$ from $\text{GNN}_{lrn}$\\
$\boldsymbol{\kappa}_{c_i}$ & KC-specific time-aware kernel for forgetting $c_i$\\
$\text{GNN}_{fgt}$& KC relation-based GNN to get parameters of $\boldsymbol{\theta}_{c_i}$\\
$\tilde{\boldsymbol{\theta}}_{c_i}^{(0)}$ & input feature of $c_i$ initialized as $\textbf{k}_{c_i}$ to $\text{GNN}_{fgt}$\\
$\tilde{\boldsymbol{\theta}}_{c_i}^{(L)}$ & output parameters of $\boldsymbol{\kappa}_{c_i}$ for $c_i$ from $\text{GNN}_{fgt}$\\

\hline
\end{tabular}
\vspace{-1em}
\end{table}

\section{Notation Table}\label{ap:notation}
We list and explain the notations in our methodology introduction in Table~\ref{tab:notation}~and~\ref{tab:notation2}.

\section{Method Details}\label{ap:graph_construction}
\subsection{KC Relation Graph Construction}
In the absence of KC relation annotations in the datasets, we construct the KC relation graph based on data statistics. For the similarity between KCs $c_i$ and $c_j$, we estimate their similarity score using:
\begin{equation*}
sim_{c_i,c_j} = \frac{\sum_{u\in\Uset}\sum_{r_{t}^u,r_{t'}^u\in\Hset^u}I(a^u_{t}=a^u_{t'},c^u_{t}=c_i,c^u_{t'}=c_j)}{\sum_{u\in\Uset}\sum_{r_{t}^u,r_{t'}^u\in\Hset^u}I(c^u_{t}=c_i,c^u_{t'}=c_j)},
\end{equation*}
where $I(\cdot)$ is the indicator function that takes value 1 if the condition is satisfied.
This approximates the probability that a student could answer questions of $c_i$ correctly while he/her could also answer questions of $c_j$ correctly (or both incorrectly), indicating an underlying similarity between them.

For the prerequisite relationship between $c_i$ and $c_j$, we assume that if $c_i$ is prerequisite to $c_j$, then answering questions of $c_i$ correctly but $c_j$ incorrectly is more likely than answering questions of $c_i$ incorrectly but $c_j$ correctly. Therefore, we use:
\begin{equation*}
pre_{c_i,c_j} = \frac{\sum_{u\in\Uset}\sum_{r_{t}^u,r_{t'}^u\in\Hset^u}I(a^u_{t}=1,a^u_{t'}=0,c^u_{t}=c_i,c^u_{t'}=c_j)}{\sum_{u\in\Uset}\sum_{r_{t}^u,r_{t'}^u\in\Hset^u}I(a^u_{t}\neq a^u_{t'},c^u_{t}=c_i,c^u_{t'}=c_j)},
\end{equation*}
to approximate the probability that $c_i$ is a prerequisite to $c_j$.

Finally, we set a threshold $\eta$ to determine whether $c_i$ is similar/prerequisite to $c_j$ (by $sim_{c_i,c_j}\geq\eta$ and $pre_{c_i,c_j}\geq\eta$, respectively). Additionally, KC pairs with a co-occurrence frequency under 10 times in the dataset are not considered.

\begin{table}[t]
\renewcommand{\arraystretch}{0.8}
\caption{Hyperparameter setting of GRKT applying for the three datasets.}
\label{tab:hyper}
\begin{tabular}{c|ccc}
\hline
Parameter & ASSIST09 & ASSIST12 & Junyi \\ \hline
$lr$      & 5e-3     & 5e-3     & 5e-3  \\
$L$       & 2        & 2        & 2     \\
$d_k$     & 16       & 16       & 16    \\
$\eta$    & 0.6      & 0.7      & 0.8   \\
$l_2$     & 1e-6     & 1e-5     & 1e-5  \\ \hline
\end{tabular}
\vspace{-1.1em}
\end{table}

\section{Supplements for Experiments}

\subsection{Metrics for Reasonability}\label{ap:reasonability_formula}
We formulate the three metrics for model reasonability in this section:
\subsubsection{Consistency}
This metric measures the ratio of consistent variation between the mastery of KCs:
\begin{equation}
\label{eq:cons}
Consistency = \sum_{u\in\Uset}\sum_{r_t^u\in\Hset^u}\frac{\sum_{c_i\in\Cset}I(m_{c_i,t}^u\geq m_{c_i,t+1}^u)}{\sum_{c_i\in\Cset}I(m_{c^u_t,t}^u\geq m_{c^u_t,t+1}^u)}.
\end{equation}
Here, we omit the averaging operation over the students and responses for conciseness. We only consider the situation where a student's mastery of the learnt KC of the current question declines while other KCs do not increase, instead of the current one increasing and the others declining. This is because the latter case might be due to natural forgetting behaviors.

\subsubsection{GAUCM}
This metric calculates the average AUC scores with respect to the mastery of each question's examined KC:
\begin{equation}
\label{eq:gauc}
GAUCM = \frac{\sum_{q_i\in\Qset}N(q_i)\cdot AUC\left[\{\hat{m}^u_{c^u_t,t}\}, \{a^u_t\}\right]^{q^u_t=q_i}_ {u\in\Uset,r_t^u\in\Hset^u}}{\sum_{q_i\in\Qset}N(q_i)}.
\end{equation}
$AUC[\hat{\mathcal{Y}}, \mathcal{Y}]_A^B$ indicates the AUC score of the prediction set $\hat{\mathcal{Y}}$ and the ground-truth set $\mathcal{Y}$, given the range $A$ and the condition $B$. 
$N(q_i)$ is the number of $q_i$ being answered.
For evaluating GRKT, we use the aggregated mastery instead of the single KC's mastery to calculate AUC because we consider the transfer of learning theory that students may leverage related KCs to solve questions.

\subsubsection{Repetition}
This metric supposes that a reasonable KT method should adhere to the following rule: after a student has finished a question and is given the same question again, the response result (correct or incorrect) should remain the same:
\begin{equation}
\label{eq:rpt}
Repetition = ACC\left[\{\textbf{KT}(q^u_t|\{r_{t'}^u|1\leq t'\leq t\})\},\{a^u_t\}\right]_{u\in\Uset,r_t^u\in\Hset^u}.
\end{equation}
$ACC(\cdot)$ denotes the accuracy score whose notation is similar to the $AUC(\cdot)$ in Equation~\ref{eq:gauc}.
$\textbf{KT}(q^u_t|\{r_{t'}^u|1\leq t'\leq t\})$ denotes the prediction score if $u$ could correctly answer $q^u_t$ given his/her past $t$ responses $\{r_{t'}^u|1\leq t'\leq t\}$ including the response to $q_t^u$ itself.

\subsection{Hyper-parameter Setting} \label{ap:hyper_setting}
We provide the hyper-parameter settings in Table~\ref{tab:hyper}. The notations on the left side indicate the learning rate, the number of GNN layers, the number of knowledge memory dimensions, the graph construction threshold, and the value of $l_2$ normalization.

\begin{table}[t]
\setlength{\tabcolsep}{3.8pt}
\renewcommand{\arraystretch}{0.9}
\caption{Comparison of GRKT applied to the Junyi dataset with labeled KC relations (GRKT-L), statistics-based relations (GRKT-S), and no relations (GRKT-0). The two values in the ``sparsity'' column respectively denote the constructed KC similarity and prerequisite graphs' sparsity.}
\label{tab:label_graph}
\begin{tabular}{l|c|ccccc}
\hline
Model  & Sparsity     & AUC    & ACC    & CONS   & GAUC   & RPT    \\ \hline
GRKT-S & 0.171, 0.169 & 0.8207 & 0.8624 & 1.0000 & 0.6473 & 0.8957 \\
GRKT-L & 0.006, 0.003 & 0.8108 & 0.8562 & 1.0000 & 0.6423 & 0.8861 \\
GRKT-0 & 0.000, 0.000 & 0.7921 & 0.8481 & 1.0000 & 0.6084 & 0.8781 \\ \hline
\end{tabular}
\vspace{-0.5em}
\end{table}

\subsection{Experimental Results Using labeled Graph Relations}\label{ap:junyi}
The Junyi dataset includes KC similarity and prerequisite relations annotated by experts with confidence scores ranging from 1 to 9.
We select relations with average scores higher than 5 as graph edges.
Table~\ref{tab:label_graph} presents the experimental results of GRKT leveraging expert-labeled relations compared with statistics-based relations and no relations.
As shown, the graphs established on expert annotations are too sparse, with only an average of 1-2 related KCs for one KC, which may not reflect real scenarios.
Despite the experimental results based on expert-labeled relations being inferior to the statistics-based version, they still exhibit noticeable improvement compared to the version without any relations. 

\begin{figure}[!t]
  \centering
  \includegraphics[width=\linewidth]{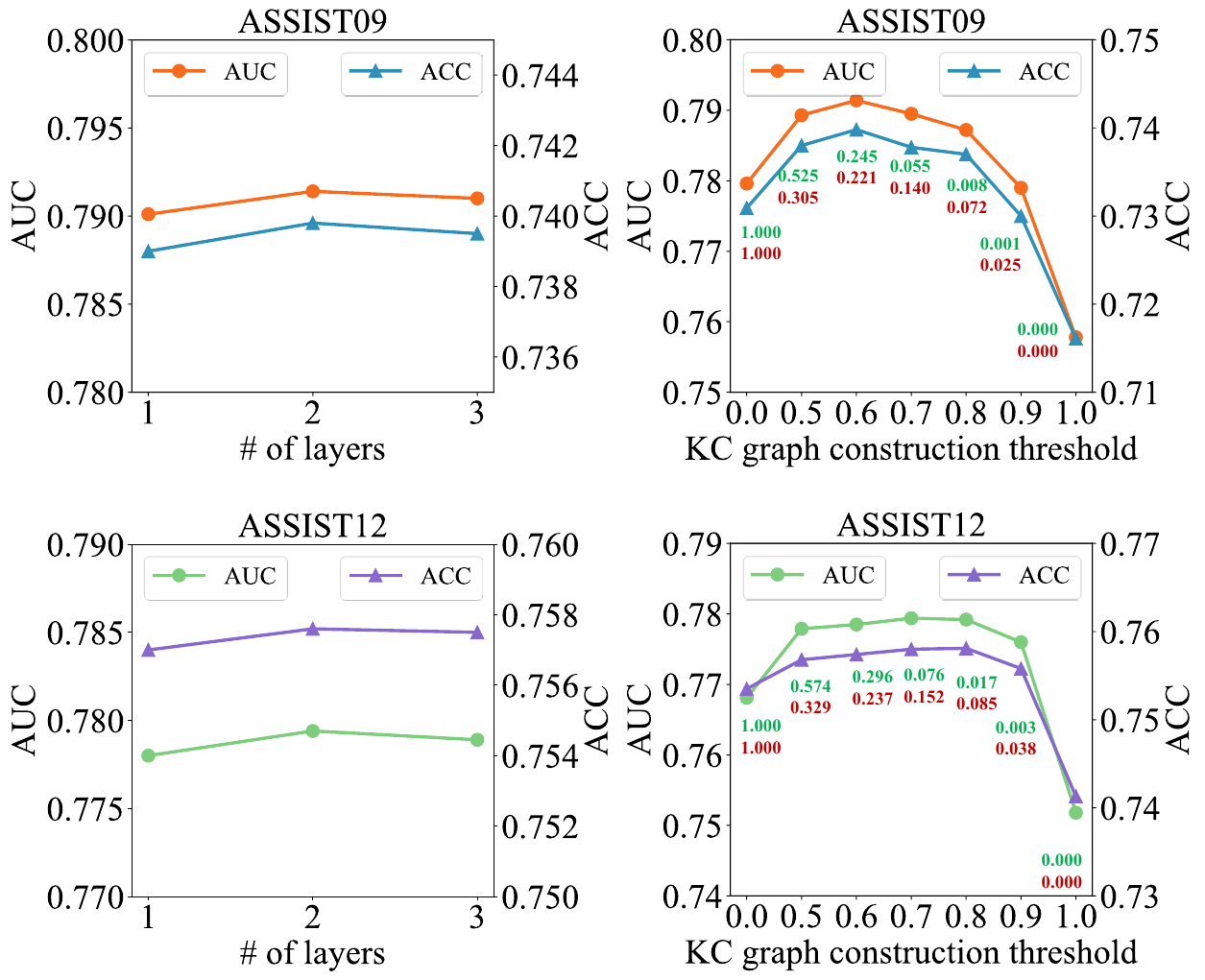}
  \caption{Experimental results analyzing the effects of hyper-parameters in GRKT are presented. The green and red decimals on the right side respectively indicate the sparsity of the constructed KC similarity and prerequisite graphs based on the specified threshold.}
 \label{fig:hyper}
 \vspace{-1.3em}
\end{figure}

\subsection{Hyper-parameter Analysis}\label{ap:hyper_analysis}
We conduct experiments to analyze the effects of various hyperparameters on GRKT's performance. The experiments are performed on the two ASSIST datasets, as shown in Figure~\ref{fig:hyper}. 
The results show that setting the number of layers in the KC relation-based graphs to 2 achieves the best performance for GRKT, suggesting that retrieving information from further distances over the graph can enhance the model. However, employing more layers may lead to overfitting issues. 
For the KC graph construction threshold, the performance peaks at around 0.6 to 0.8. In this interval, the sparsity of the two graphs ranges from 0.01 to 0.3, indicating that too many relations lead to structural redundancy, while too few result in limited information sharing between KCs.

\balance

\end{document}